\documentclass{article}
\usepackage{ijcai21}

\usepackage{times}

\usepackage{soul}
\usepackage{verbatim}
\usepackage{url}
\usepackage{multirow}
\usepackage[hidelinks]{hyperref}
\usepackage[utf8]{inputenc}
\usepackage[small]{caption}
\usepackage{graphicx}
\usepackage{amsmath}
\usepackage{booktabs}
\usepackage{mathbbol}
\title{Learning Class-Transductive Intent Representations for \\ Zero-shot Intent Detection}

\author{
Qingyi Si$^{1,2}$\and
Yuanxin Liu$^{1,2}$\and
Peng Fu$^1$\footnote{Peng Fu and Zheng Lin are the corresponding authors.}\and
Zheng Lin$^{1*}$\and
Jiangnan Li$^{1,2}$\and
Weiping Wang$^1$\\
\affiliations
$^1$
Institute of Information Engineering, Chinese Academy of Sciences, Beijing, China\\
$^2$School of Cyber Security, University of Chinese Academy of Sciences, Beijing, China\\
\emails
\{siqingyi, liuyuanxin, fupeng, linzheng, lijiangnan, wangweiping\}@iie.ac.cn
}
\begin{document}

\maketitle
\begin{abstract}
Zero-shot intent detection (ZSID) aims to deal with the continuously emerging intents without annotated training data. However, existing ZSID systems suffer from two limitations: 1) They are not good at modeling the relationship between seen and unseen intents. 
2) They cannot effectively recognize unseen intents under the generalized intent detection (GZSID) setting. A critical problem behind these limitations is that the representations of unseen intents cannot be learned in the training stage. 
To address this problem, we propose a novel framework that utilizes unseen class labels to learn \textbf{C}lass-\textbf{T}ransductive \textbf{I}ntent \textbf{R}epresentations (CTIR). Specifically, we allow the model to predict unseen intents during training, with the corresponding label names serving as input utterances. 
On this basis, we introduce a multi-task learning objective, which encourages the model to learn the distinctions among intents, and a similarity scorer, which estimates the connections among intents more accurately. 
CTIR is easy to implement and can be integrated with existing methods. 
Experiments on two real-world datasets show that CTIR 
brings considerable improvement to the baseline systems.\footnote{The code, datasets and Appendix are available at \url{https://github.com/PhoebusSi/CTIR}}
\end{abstract}

\section{Introduction}
\label{sec:intro}
In recent years, smart devices with built-in personal assistants like Google Assistant and Siri are becoming omnipresent. Behind these systems, a key question is how to identify the underlying intent of a user utterance, which has triggered a large amount of work on intent detection \cite{ravuri2015recurrent,LiuL16,nam2016all}. Most existing intent detection systems are built on models trained on annotated data. However, as user demands and the functions of smart devices continue to grow, collecting supervised data for every new intent becomes labor-intensive.

To address this issue, some studies tackle intent detection in the zero-shot learning (ZSL) manner, attempting to utilize the learned knowledge of seen classes to help detect unseen classes. Recent methods of ZSID can be roughly divided into two categories: The first category \cite{XiaZYCY18,liu2019reconstructing}, referred to as the transformation-based methods, utilizes word embeddings of label names to establish a similarity matrix, which is then used to transfer the prediction space of seen intents to unseen intents. Another line is the compatibility-based methods \cite{chen2016zero,kumar2017zero}, which aims to encode the label names and utterances into the same semantic space and then calculate their similarity. However, in both kinds of methods, most existing ZSID methods are \textit{inductive}, which do not consider any information about the unseen classes in the training stage. Consequently, the representations of unseen intents cannot be learned, resulting in two limitations.

First, the ZSID methods are not good at modeling the relationship between seen and unseen intents, especially when the label names are given in the form of raw phrases or sentences. For the transformation-based methods, word embeddings of label names are inadequate to associate the connections between seen and unseen intents. For example, “BookRestaurant” is similar to “RateBook” when measured by word embeddings. However, the meaning of these two intents are not that relevant.
For the compatibility-based methods, since the unseen intents are not included in learning the semantic space shared by utterance and label names, it cannot effectively detect unseen intents during the test stage, especially when the expressions of utterances are diverse.

Second, the vanilla ZSL methods are not applicable to generalized zero-shot intent detection (GZSID), where the models (at test time) are presented with not only unseen class utterances but also seen class utterances.
In GZSID, existing ZSL models usually suffer from the domain shift \cite{fu2015transductive} problem, in which utterances from unseen intents are almost always mistakenly classified into seen intents.

The two limitations are caused by inadequate learning of unseen intents. Naturally, the label name provides a proper sketch of the intent meaning. Existing models use it during the test stage. In contrast, we introduce the \textit{class-transductive} \cite{ZSL_survey,Yongqin19} setting into ZSID, which uses semantic information about the unseen classes (e.g., the label names) for model training. 
Specifically, we include the unseen intents into the prediction space during training, with the label names serving as the pseudo utterances. 
This allows the model to learn a rough boundary of each seen and unseen class in the semantic space.
Under this framework, we introduce an assistant task that forces the model to find the distinction between seen and unseen intents, thereby alleviating the domain-shift problem. On this basis, we refine the word embedding based similarity matrix by averaging the representations of all corresponding (seen intent) utterances and (unseen intent) label names. As a result, we can better capture the intent meanings and the similarity matrix reflects more accurate intent connections.
In summary, our contribution is three-fold:
\begin{itemize}

\setlength{\itemsep}{0pt}
\setlength{\parsep}{0pt}
\setlength{\parskip}{0pt}
\item{In response to the limitations of existing ZSID systems, we propose a class-transductive framework that makes use of unseen label names in the training stage.
}

\item{Under the framework, we present a multi-task learning objective to find the inter-intent distinctions and a similarity scorer to associate the inter-intent connections.}

\item{Empirical results on ZSID and GZSID in two benchmarks show that CTIR can bring improvement to a wide range of ZSID systems with different zero-shot learning strategies and model architectures.}
\end{itemize}

\begin{figure}[t]
\centering
\includegraphics[width=0.9\linewidth]{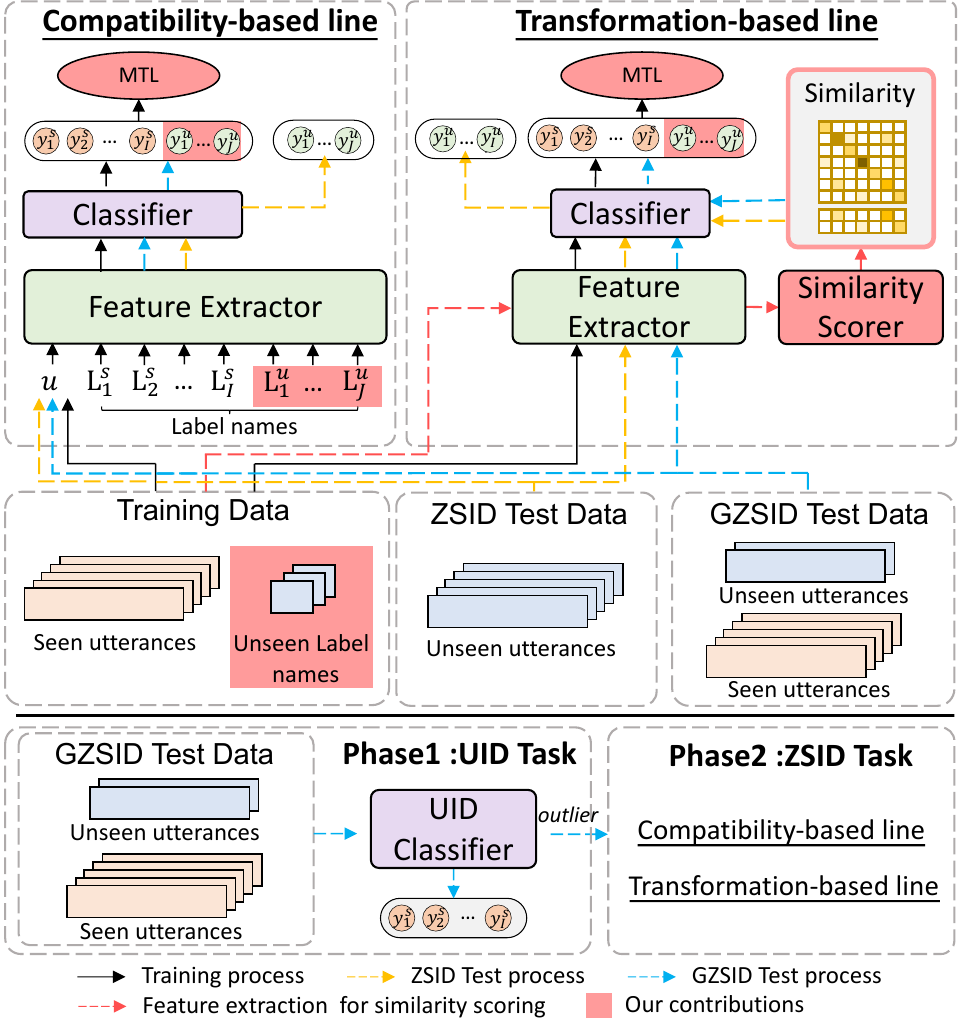} 
\caption{Illustration of how to integrate compatibility-based method (upper left), transformation-based method (upper right) and two-stage GZSID method (lower) with CTIR.}
\label{fig:architecture}
\vspace{-0.2cm}
\end{figure}
\section{Problem Formulation}
\label{sec:problem_formulate}
\paragraph{Zero-shot Intent Detection.}
In ZSID, the model is trained on the annotated dataset \{($x$, $y$)\}, where $y\,\in\,Y_{seen}=\,\{y^{s}_1, y^{s}_2, \dots, y^{s}_I\}$ is the intent label and $x$ is the utterance. At test time, the goal is to identify the intent of an utterance, which belongs to one of the $J$ unseen intents $Y_{unseen}=\{y^{u}_{1}, y^{u}_{2}, \dots, y^{u}_{J}\}$, where $Y_{seen}\cap Y_{unseen}=\emptyset$. 

\paragraph{Generalized Zero-shot Intent Detection.}
In GZSID, the model is presented with utterances from either seen or unseen intents, and the prediction space is $Y_{seen} \cup Y_{unseen}$.
\paragraph{Unknown Intent Detection (UID).} 
In UID~\cite{LinX19}, the training set is the same as ZSID and GZSID. During testing, the model is expected to detect seen intents and decide whether an utterance belongs to the unknown intents. The prediction space is $\{y^{s}_1, y^{s}_2,\dots, y^{s}_I, y_{unseen}\}$, where all the unknown intents are grouped into a single class $y_{unseen}$. The GZSID can be solved by breaking it into UID and ZSID tasks(as shown in Figure1 lower).

\paragraph{Simplified Unknown Intent Detection (SUID).} The prediction space is reduced to $\{y_{seen}, y_{unseen}\}$ in SUID. In our multi-task learning, SUID serves as an assistant task.

\section{CTIR Framework}
In this section, we describe how to integrate CTIR into transformation-based, compatibility-based methods and  two-stage GZSID method respectively. 
Figure\ref{fig:architecture} gives an overview. The core idea is to expand the prediction space during training to include unseen classes, with the unseen label names serving as pseudo utterances. During test time, the trained model can be applied to both ZSID and GZSID settings.

\subsection{Transformation-based Methods}
\label{sec:tranform-based}
\paragraph{Feature Extractor.}
The feature extractor transforms an input text into a sequence of hidden vectors $\mathbf{H}=(\mathbf{h}_1,\mathbf{h}_2,\dots,\mathbf{h}_{T})\in \mathbb{R}^{T\times D_H}$, where $D_H$ is the hidden dimension and $T$ is the sequence length. 
Note that the architecture of feature extractor is a free choice in CTIR framework. Specially,  on top of the $\mathbf{H}$ produced by a Bi-LSTM, 
CapsNet \cite{XiaZYCY18} applies a multi-head attention layer. Each attention head represents a unique semantic feature, which gives rise to $\{\mathbf{m}_{1}, \mathbf{m}_{2}, \dots \mathbf{m}_{R}\} \in \mathbb{R}^{R\times D_H}$, where $R$ is the number of attention heads. We follow this operation when combining CapsNet with CTIR.
\paragraph{Intent Classifier.}
There are two commonly used classifiers in intent detection: 1) the linear classifier and 2) the capsule networks \cite{sabour2017dynamic}. The linear classifier predicts the intent probabilities as:
\begin{equation}
  \mathbf{v}_{tr}=\mathrm{Softmax}((\frac{1}{T} \sum_{t=1:T} \mathbf{h}_{t}) \mathbf{W})
\end{equation}
where $W\in \mathbb{R}^{D_H\times K}$ is the weight matrix, and $K=I+J$ is the total number of seen and unseen classes. 

In capsule networks, the Dynamic Routing algorithm \cite{sabour2017dynamic} is used to aggregate the low-level features $\{\mathbf{m}_{1}, \mathbf{m}_{2}, \dots \mathbf{m}_{R}\}$ into higher-level representations:
\begin{equation}\begin{array}{c}
\mathbf{\hat{u}}_{k \mid r} = \mathbf{m}_{r} \mathbf{W}_{k, r} \\
\mathbf{s}_{k}=\sum_{r} c_{k r} \mathbf{\hat{u}}_{k \mid r}
\end{array}\end{equation}
where $c_{kr}$ is the coefficient that determines how much the $r^{th}$ semantic feature contributes to intent $y_k$. Following $\mathbf{s}_k$ is the squash function, which gives rise to the activation vectors $\mathbf{V}=(\mathbf{v}_1,\mathbf{v}_2,\dots,\mathbf{v}_K)$. The 2-norms of these vectors are then used as the probability of different intent classes.

\paragraph{Multi-task Learning Objective.}
The assistant task SUID has two class labels, namely $y_{seen}$ and $y_{unseen}$.  
In order to compute the probabilities for $y_{seen}$ and $y_{unseen}$, we sum up the probabilities of all intent classes in the respective categories. For linear classifiers, we compute the sum of the first $I$ dimensions of the vector $\mathbf{v}_{tr}$ for $y_{seen}$ and the last $J$ dimensions for $y_{unseen}$. 
Then, the linear classifier is trained with cross-entropy loss:
\begin{equation}
\label{equ:multi-task1}
\mathcal{L}_{cross}=\sum_{k=1}^{K} z_{k} \log \left(p_{k}\right)+\alpha\sum_{n=1}^{2} z_{n}^{\prime} \log \left(P_{n}\right)
\end{equation}
where $z_k \in \{0, 1\}$ indicates whether the $k^{th}$ intent is true, and $p_k$ is the predicted probability of the $k^{th}$ intent. $z_{n}^{\prime}$ and $P_n$ are respectively the ground truth and the predicted probability of each class in SUID. $\alpha$ is a down-weighting coefficient.

For capsule networks, we sum up the 2-norms of the activation vectors $(\mathbf{v}_1,\mathbf{v}_2,\dots,\mathbf{v}_I)$ and $(\mathbf{v}_{I+1},\mathbf{v}_{I+2},\dots,\mathbf{v}_K)$, respectively. Then, the max-margin loss for capsule networks is:
\begin{equation}
\begin{aligned}
\mathcal{L}_{margin}=& \sum_{k=1}^{K}\left\{T_{k} \cdot \max \left(0, m^{+}-\left\|\mathbf{v}_{k}\right\|\right)^{2}\right.\\
&\left.+\lambda\left(1-T_{k}\right) \cdot \max \left(0,\left\|\mathbf{v}_{k}\right\|-m^{-}\right)^{2}\right\} \\
&+\lambda^{\prime} \sum_{n=1}^{2}\left\{T_{n} ^{\prime} \cdot \max \left(0, m^{\prime+}-P_{n}\right)^{2}\right.\\
&\left.+\lambda\left(1-T_{n}^{\prime}\right) \cdot \max \left(0, P_{n}-m^{\prime-}\right)^{2}\right\}
\end{aligned}
\end{equation}
where $T_k = 1$ when the $k^{th}$ intent is ground-truth, and otherwise $T_k = 0$. $T_{n}^{\prime}$ is defined in the same way for SUID. $\lambda$ and $\lambda^{\prime}$ are the down-weighting coefficients, $m^{+}$, $m^{-}$ and $m^{\prime+}$, $m^{\prime-}$ are the margins. In addition, a regularization term is added to $\mathcal{L}_{margin}$ to encourage the discrepancy among attention heads \cite{XiaZYCY18}.

\paragraph{Similarity Scorer.}
Similarity Scorer, which measures the connections between intent classes, is a key component for transformation-based methods. 
Inspired by Chao~\shortcite{chao2016empirical}, we average the representations of all utterances to represent the seen intents. For the unseen intents, we use the representations of label names. The representation of each utterance or label name is obtained by averaging over different time steps 
 or attention heads (for CapsNet). In practice, the representations are computed during the last training epoch, which considers the entire training set (i.e., the parameters of feature extractors are updated in this process).

After the averaging operation, we have the intent representations $\{\mathbf{g}_{1}, \mathbf{g}_{2}, \dots \mathbf{g}_{K}\} \in \mathbb{R}^{K \times D_H}$, which is used to compute the similarity matrices $\mathbf{L}_{zsl}\in \mathbb{R}^{K\times J}$ for ZSID and $\mathbf{L}_{gzsl}\in \mathbb{R}^{K\times K}$ for GZSID. The similarity between intent $k_1$ and $k_2$ is computed as ${L}_{k_1 k_2}=\mathrm{Cosine}\left(\mathbf{g}_{k_{1}},\mathbf{g}_{k_{2}}\right)$, where $\mathbf{g}_{k1}$, $\mathbf{g}_{k2}$ are the representations of $k1$ and $k2$, respectively. 
\paragraph{Inference Process.}
During inference, a test utterance is first encoded into a prediction vector of $K$ dimensions, in the same way as the training process. Based on this vector, we further employ the Similarity Scorer to obtain the final prediction. In terms of the linear classifier, we have:
\begin{equation}
    \mathbf{v}_{te}=\mathrm{Softmax}((\frac{1}{T} \sum_{t=1:T} \mathbf{h}_{t}) \mathbf{W} \mathbf{L})
\end{equation}
where $\mathbf{L}$ refers to $\mathbf{L}_{zsl}$ or $\mathbf{L}_{gzsl}$. When it comes to the capsule networks:
\begin{equation}
\mathbf{s}_{j}=\sum_{r} L_{j k}(c_{k r} \mathbf{\hat{u}}_{j | k})
\end{equation}
where $L_{j k}$ is the $k^{th}$ entry in the $j^{th}$ row of $\mathbf{L}$. After Dynamic Routing, we can get $J$ activation vectors for ZSID and $K$ activation vectors for GZSID.

\subsection{Compatibility-based Methods}
\label{sec:compati-based}
\paragraph{Feature Extractor.}
Similar to the transformation-based methods, the first step of compatibility-based methods is to encode the utterance or label name into a dense vector. In this paper we study two kinds of compatibility-based methods: Zero-shotDNN \cite{kumar2017zero} and CDSSM \cite{chen2016zero}, which extract the text representation with a tanh-activated nonlinear layer and CNN respectively. 
\paragraph{Intent Classifier.}
With the representations of an utterance and all label names, compatibility-based methods compute the cosine similarity between the utterance $u$ and each intent, which results in $\mathbf{S}=\{sim(u, y) | y \in Y_{seen} \cup Y_{unseen}\} \in \mathbb{R}^{K}$. Then, $u$ can be classified into a particular intent class according to $\mathbf{v}_{tr}=\mathrm{Softmax}(\mathbf{S})$.

\paragraph{Multi-task Learning Objective.}
To perform SUID with the compatibility-based classifier, we sum up the first $I$ and last $J$ positions in $\mathbf{S}$ as $\sum_{i=1:I} \mathbf{S}_{i}$ and $\sum_{i=I+1:K} \mathbf{S}_{i}$, the result of which is passed to a binary softmax function to obtain the final probabilities for $y_{seen}$ and $y_{unseen}$. The multi-task learning objective is defined in the same way as Equation \ref{equ:multi-task1}.

\paragraph{Inference Process.}
The inference process of compatibility-based methods is basically the same as the classification process during training. Given an input utterance, we compute its similarity with each candidate intent in the learned representation space, and classify it to the most similar intent.

\subsection{Two-stage GZSID Method}
\label{sec:two-stage}

The main idea of two-stage method is to first determine whether an utterance belongs to unseen intents (i.e., $Y_{seen}$), and then classify it into a specific intent class. This method bypasses the need to classify an input sentence among all the seen and unseen intents, thereby alleviating the domain shift problem. To verify the performance of integrating CTIR into the two-stage method, we design a new two-stage pipeline. In Phase1, a test utterance is classified into one of the classes from $Y_{seen} \cup \{y_{unseen}\}$ using the UID classifier. In practice, we use the density-based algorithm LOF (LMCL) \cite{LinX19} to perform UID. In Phase2, we perform ZSID for the utterances that have been classified into $y_{unseen}$, using the methods described in Section \ref{sec:tranform-based} and Section \ref{sec:compati-based}.

\section{Experiments}
\subsection{Datasets and Experimental Setup}
We conduct experiments on two benchmarks for intent detection.  
~\textbf{SNIPS} \cite{coucke2018snips} is a corpus to evaluate the performance of voice assistants, which contains 5 seen intents and 2 unseen intents. \textbf{CLINC} \cite{Stefan} includes out-of-scope queries and 22,500 in-scope queries covering intent classes from 10 domains. We use the in-scope data to build our dataset with 50 seen intents and 10 unseen intents. For more details of the datasets, please refer to Appendix A$^1$. 
\paragraph{Dataset Processing.} For ZSID, we use all utterances from seen intents to construct the training set and those from unseen intents to construct the test set. For the training set of GZSID, we randomly select 70\% utterances of each seen intent and replicate the unseen label names to roughly the same number. Although our focus is the zero-shot problem, utterances from seen intents still account for the majority in real-world applications. In light of this phenomenon, we balance the sample number of unseen and seen classes to build the test set: selecting the remaining 30\% utterances of each seen intent and 30\% random utterances of each unseen intent ($30\%$ seen and $100\%$ unseen intent utterance in the setting of existing works). 
Besides, we use the raw label names without modification while most existing work modified the label names slightly to achieve better performance, e.g. from ``SearchScreeningEvent" to ``SearchMovie".

\paragraph{Experimental Setup.}
We use the test set for hyper-parameter(Appendix B$^1$) tuning, which is the same with most ZSID work. 
We run the experiments for five times with different random seeds. 
We consider two evaluation metrics: Accuray (Acc), 
and F1, both of which are computed with the average value weighted by their support on each class. 

\subsection{Baselines}
We integrate CTIR with representative ZSID systems. For transformation-based methods with linear classifier, we explore the use of CNN, LSTM, and BERT \cite{BERT} as the feature extractor, which are denoted as CNN, LSTM and BERT, respectively. We find that use BERT to compute the similarity matrix leads to poor results, as the label names are very short. Therefore, we combine BERT feature extractor with CNN-CTIR computed similarity matrix for the BERT baseline. For capsule network classifier, we adopt CapsNet \cite{XiaZYCY18} as the baseline system. In terms of the compatibility-based methods, we study the combination of CTIR with Zero-shotDNN \cite{kumar2017zero} and CDSSM \cite{chen2016zero}.
\begin{table}[t]
\begin{center}
\resizebox{0.9\linewidth}{!}{
\begin{tabular}{l|cc|cc}
\hline
\multirow{2}{*}{Model}      & \multicolumn{2}{c}{SNIPS}  & \multicolumn{2}{|c}{CLINC}                \\ \cline{2-5} 
&Acc  &F1   &Acc    &F1      \\ \hline

LSTM (Ours)  &79.47 &79.18  &71.73 &68.73\\
\quad +CTIR  &\bf93.43 &\bf93.42  &\bf84.48 &\bf84.35 \\ \hline

CNN (Ours)   &65.15 &60.91  &73.03 &70.94   \\
\quad +CTIR  &\bf94.73 &\bf94.73  &\bf85.11 &\bf85.20\\ \hline

BERT (Ours) &73.66 &73.32  &62.73 &59.45\\
\quad +CTIR  &\bf96.13 &\bf96.12  &\bf88.72 &\bf88.45\\ \hline

CDSSM (Chen et al. \citeyear{chen2016zero})  &68.98 &66.52  &64.80 &61.14     \\
\quad +key   &83.03 &82.86 &- &- \\
\quad +CTIR   &\bf94.14 &\bf94.14  &\bf83.07 &\bf82.52\\ \hline

ZSDNN \cite{kumar2017zero} &80.96 &80.74  &82.20 &82.18 \\
\quad +key    &93.49 &93.49  &- &-\\
\quad +CTIR    &\bf95.07 &\bf95.07  &\bf93.57 &\bf93.62\\ \hline

CapsNet \cite{XiaZYCY18}  &74.21 &72.58  &64.51 &61.87  \\
\quad +key  &93.49 &93.49 &- &-\\
\quad +CTIR  &\bf94.84 &\bf94.84 &\bf87.01 &\bf86.91\\ \hline
\end{tabular}

}
\caption{Results of ZSID. (Ours) represents our implementation.}
\label{tab_ZSID}
\end{center}
\end{table}
\begin{table*}[t]
\begin{center}
\resizebox{0.85\textwidth}{!}{
\begin{tabular}{l|cc|cc|cc|cc|cc|cc}
\hline
\multirow{3}{*}{Model}      & \multicolumn{6}{c}{SNIPS}  & \multicolumn{6}{|c}{CLINC}                  \\ \cline{2-13}&

\multicolumn{2}{c}{Seen}  & \multicolumn{2}{|c}{Unseen} & \multicolumn{2}{|c}{Overall} & \multicolumn{2}{|c}{Seen}  & \multicolumn{2}{|c}{Unseen} & \multicolumn{2}{|c}{Overall}                  \\ \cline{2-13}

&Acc  &F1   &Acc  &F1   &Acc  &F1   &Acc  &F1   &Acc  &F1   &Acc  & F1      \\ \hline

LSTM (Ours) &\bf97.22 &83.94 &0.00 &0.00  &69.66 &60.14 &93.19 &85.95 &0.00 &0.00	&77.65 &71.62 \\
\quad +CTIR  &95.42 &\bf86.00 &\bf31.34 &\bf45.32  &\bf77.26 &\bf74.47 &\bf93.40 &\bf87.55 &\bf20.50 &\bf30.96 &\bf81.24 &\bf78.12\\ \hline

CNN (Ours)  &\bf97.95 &85.09 &0.00 &0.00  &70.18	&60.97 &95.35 &87.89 &0.00 &0.00 	&79.46	&73.24\\ 
\quad +CTIR  &96.59 &\bf88.41 &\bf46.22 &\bf62.37 &\bf82.31 &\bf81.03 &\bf96.25 &\bf90.06 &\bf15.82 &\bf23.11  &\bf82.85 &\bf78.91\\ \hline

BERT (Ours)  &96.43 &84.70 &0.00 &0.00	&69.1	&60.69 &88.14 &82.17 &0.23 &0.44 	&73.49 &68.55\\ 
\quad +CTIR  &\bf96.93 &\bf89.62 &\bf49.38 &\bf61.15  &\bf83.45 &\bf81.55 &\bf97.84 &\bf91.19 &\bf12.55 &\bf19.74 &\bf83.62 &\bf79.29\\ \hline
CDSSM (Chen et al. \citeyear{chen2016zero})     &\bf97.99 &84.54 &0.19 &0.37  &70.32 &60.74 &93.62 &86.66 &2.09 &3.69  &78.35 &72.82 \\
\quad +LOF &81.52 &\bf88.22 &60.75 &49.14  &75.63 &77.14 &79.96 &86.94 &42.80 &28.96 &73.76 &77.28 \\
\quad +CTIR &92.55 &84.63 &42.91 &56.46  &78.51	&76.66  &\bf94.72  &\bf89.11 &23.64 &32.39 &\bf82.87 &79.65 \\
\quad +LOF+CTIR   &81.52	&\bf88.22 &\bf87.00  &\bf73.01	&\bf83.08	&\bf83.90 &79.96	&86.94  &\bf61.32	&\bf44.58  &76.89	&\bf79.97\\ \hline

ZSDNN \cite{kumar2017zero} &94.79 &83.29 &10.50 &17.40 &70.93 &64.64 &85.45 &81.14 &26.79 &36.07  &75.63 &73.73\\ 
\quad +LOF  &81.52 &88.22 &66.37 &55.24 &77.23 &78.86 &79.96 &86.94 &64.04 &47.13 &77.30 &80.30\\
\quad +CTIR  &\bf95.82 &\bf89.11 &58.03 &73.00 &\bf85.12 &\bf84.55 &\bf91.75 &\bf88.18 &47.15 &\bf58.69 &\bf84.30 &\bf83.18\\ 
\quad +LOF+CTIR &81.52	&88.22 &\bf88.52 &\bf74.30 &83.51	&84.27 &79.96	&86.94  &\bf75.47	&54.80 &79.21  &81.85
\\ \hline

CapsNet \cite{XiaZYCY18} &\bf97.99 &84.34 &0.00 &0.00  &70.22 &60.43 &\bf97.92 &90.31 &0.23 &0.42  &81.64 &75.33 \\
\quad +LOF  &81.52 &88.22 &64.74 &53.00	&76.76 &78.23 &79.96 &86.94 &46.04 &33.80  &74.30 &78.08 \\
\quad +CTIR  &97.47 &\bf89.15 &47.17 &63.90	&\bf83.21 &81.99 &97.71 &\bf92.29 &30.95 &43.09  &\bf86.58 &\bf84.09 \\ 
\quad +LOF+CTIR &81.52	&88.22  &\bf86.63	&\bf72.91  &82.97	&\bf83.88   &79.96	 &86.94  &\bf64.40	&\bf46.76  &77.36	&80.24\\ \hline
\end{tabular}
}
\caption{Results of GZSID. ZSDNN is short for Zero-shotDNN.}
\label{tab_GZSID}
\end{center}
\end{table*}

We also introduce two strong baselines targeting the two limitations related to ZSID and GZSID respectively, as discussed in Section \ref{sec:intro}. For ZSID, we replace the original label names with manually selected keywords (denoted as + key), which reflect the inter-intent relationships more accurately.
For GZSID, we build a two-stage approach (denoted as + LOF) as described in Section \ref{sec:two-stage}.

ReCapsNet \cite{liu2019reconstructing} and the two-stage method SEG \cite{Guangfeng}, 
which achieve the SOTA performance, are not open-sourced. 
Different from our two-stage method, SEG ensembles UID and ZSID models in Phase1 and conducts seen intent prediction in Phase2. 
To compare with them, we follow their way of data processing and run our model.

\subsection{Results and Analysis}
\paragraph{Performance in ZSID.}
As can be seen in Table \ref{tab_ZSID}, the performance of CDSSM, ZSDNN and CapsNet can be significantly improved with the manually composed keyword labels. This demonstrates the first limitation that existing ZSID methods are not good at associating the relationship between seen and unseen intents when the label names are given in raw phrases or sentences. Our proposed framework, regardless of the backbone network and ZSID strategy, consistently outperforms the baselines with comfortable margin (11.49 $\sim$ 33.82 absolute F1 improvement). More importantly, the CTIR-enhanced models also achieve better results than the manually selected keywords, with an encouraging improvement of 11.28 F1 score for the CDSSM baseline. This shows the effectiveness of the proposed method to alleviate the effect of poor-quality label names, dispensing with the need of manual modification.

\paragraph{Performance in GZSID.}
From the results in Table \ref{tab_GZSID}, we can derive four observations: 1) The baselines work well on the seen intents, while they fail to detect the unseen intents. This phenomenon attests to the second limitation that the ZSID systems cannot effectively work in the GZSID scenario. 2) The two-stage framework (+LOF) brings significant improvement on the unseen intents, successfully alleviating the domain-shift problem. 3) Our CTIR framework improves the performance to a larger extent, which on average outperforms +LOF with 10.04 F1 on unseen intents and 3.37 F1 in terms of overall performance.
4) In terms of Acc, CTIR performs better in seen intents while LOF achieves higher scores in unseen intents. We analyze this trade-off phenomenon in Appendix C$^1$.
5) CTIR can further promote the strong two-stage baseline: Enhancing the Phase2 ZSID model with CTIR (+LOF+CTIR) results in a substantial improvement of +LOF on the unseen intents, where the Acc is increased by 19.77 on average and F1 score is increased by 16.52 on average. The overall performance is thereby also improved.

\begin{figure}[ht]
\centering
\includegraphics[width=1.0\linewidth]{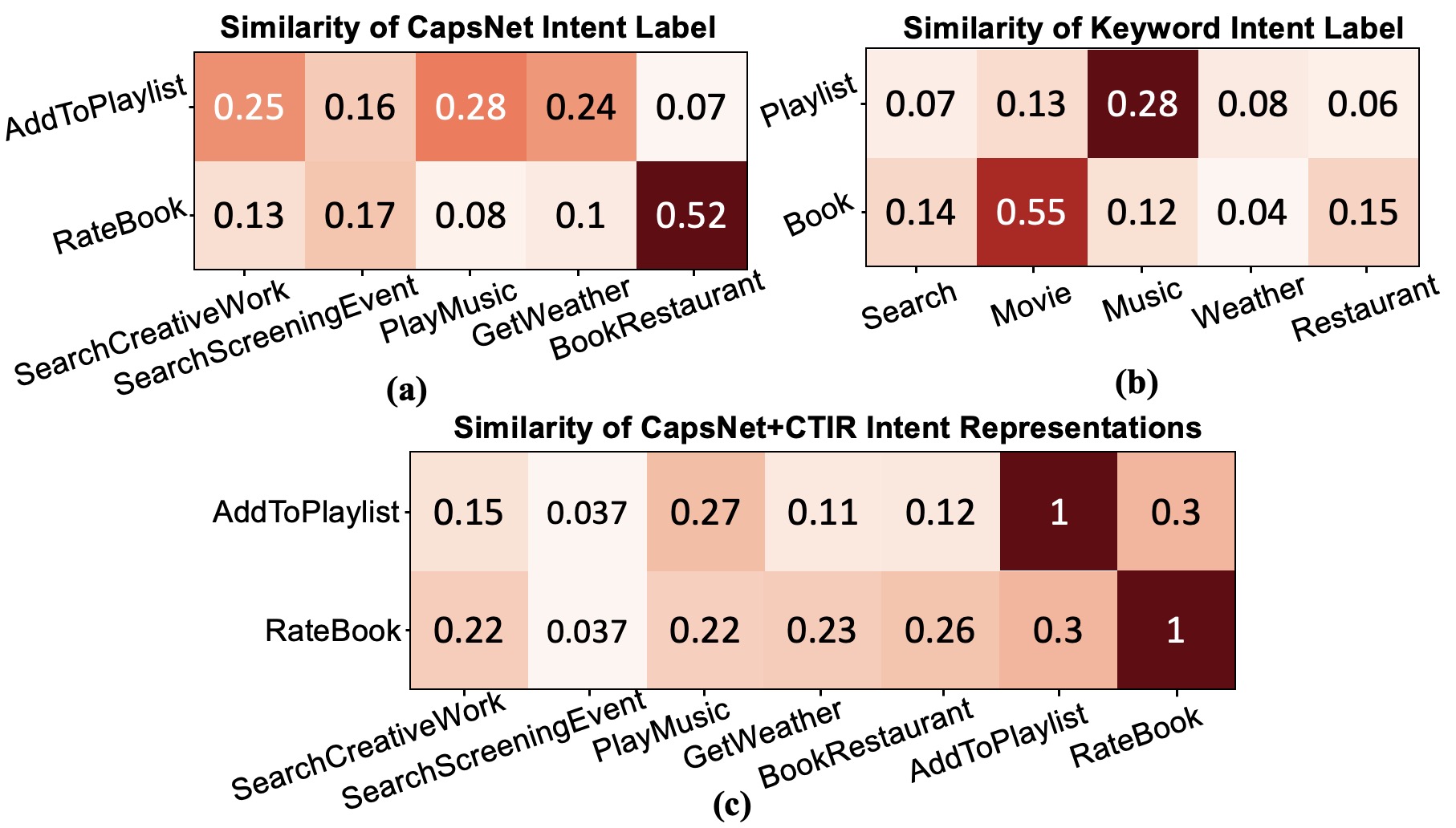} 
\caption{The similarities computed by (a) CapsNet, (b) keyword intent labels and (c) CapsNet+CTIR. Note that the similarity scores for CapsNet+CTIR are not normalized in our experiments.}
\label{fig:similarity}
\vspace{-0.1cm} 
\end{figure}

\begin{figure}[ht]
\centering
\includegraphics[width=1.0\linewidth]{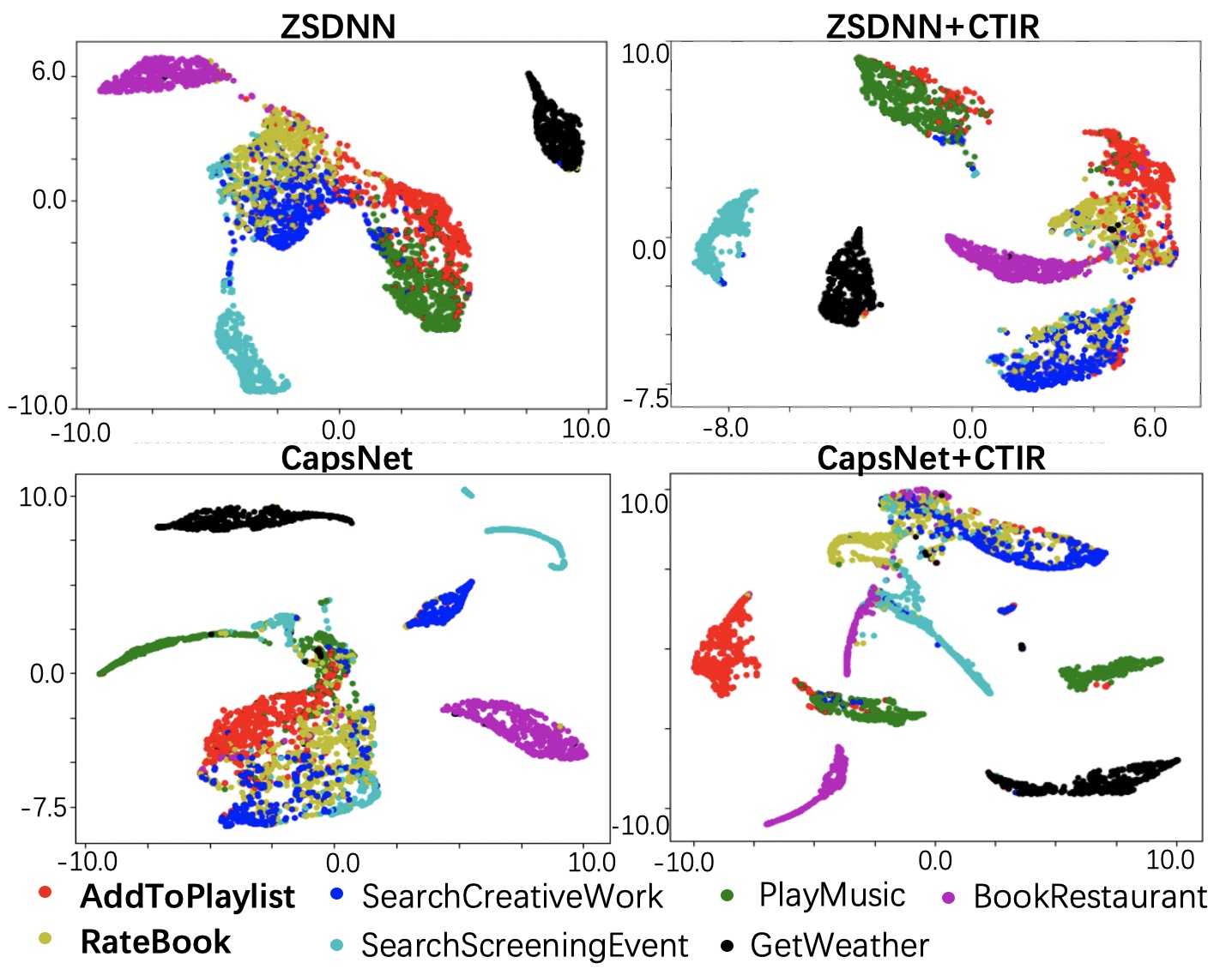} 
\caption{Data visualization on SNIP test set. ``AddToPlaylist" and ``RateBook" are the unseen intents.}
\label{fig:vectorspace}
\vspace{-0.1cm}
\end{figure}

\paragraph{Relieving of two limitations.}
We qualitatively analyze the effectiveness of CTIR in relieving the two limitations by answering the following questions. Q1: Can CTIR model a better inter-intent relationship in ZSID? Q2: Can CTIR distinguish between the samples of seen and unseen intents?

~\textbf{Q1}
A good inter-intent relationship is crucial to transform the model from seen intent to unseen intent.
To illustrate the problem of word embedding based inter-intent similarity, we compute the similarity with the sum of original label word embeddings (denoted as CapsNet Intent Labels), the keyword embedding (denoted as Keyword Intent Labels), and our similarity scorer (CapsNet+CTIR), which is shown in Figure\ref{fig:similarity}. 
When using the original label names, ``BookRestaurant" is much more similar to ``RateBook" than the other labels, as a result of the shared word ``Book". In comparison, the keyword-based similarity matrix and our similarity scorer successfully avoid the pitfall of ``Book" polysemy, and the similarities among other intents are also reasonable (e.g., ``AddToPlaylist" and ``PlayMusic"). This demonstrates that the intent representation of CapsNet+CTIR is well-learned, which can better associate the inter-intent relationship and dispenses with manually selecting keywords.
We also find that for CapsNet+CTIR the two unseen classes are more similar to each other than to the seen classes. This is partly due to the introduction of the simplified unseen intent detection task, which aims to distinguish between seen classes and unseen classes.

~\textbf{Q2} 
To examine whether the CTIR framework can actually learn to distinguish between the unseen and seen intent representations, we visualize the utterance representations using the feature extractors of four methods (see Figure \ref{fig:vectorspace}). We can see that the two unseen intents ``AddToPlaylist" (red) and ``RateBook" (yellow) are entangled with ``SearchCreativeWork" (blue) and ``PlayMusic" (green) in the CapsNet representation space. For ZSDNN representations, yellow dots and blue dots appear as a single cluster, so it is with red dots and green dots. These entangled representations make it difficult to perform GZSID. In comparison, the CapsNet+CTIR successfully learns an independent sub-space for ``AddToPlaylist" and disentangles a part of ``RateBook" from the seen intents. ZSDNN+CTIR pulls the two unseen classes as a whole out from the seen classes, with a vague but identifiable boundry in between.

\begin{table}[t]
\begin{center}
\resizebox{0.9\linewidth}{!}{
\begin{tabular}{l|cc|cc}
\hline
\multirow{2}{*}{Model}      & \multicolumn{2}{c}{ZSID}  & \multicolumn{2}{|c}{GZSID}                  \\ \cline{2-5} 
&Acc  &F1   &Acc   &F1 \\\hline

ReCapsNet \cite{liu2019reconstructing}  &79.96	&79.80	&47.05	&38.26 \\
SEG \cite{Guangfeng} &-	&-	&76.85	&76.74  \\ \hline

ZSDNN+CTIR &\bf95.05	&\bf95.05 &74.41	&75.44  \\
ZSDNN+LOF  &-  &-  &76.93	&77.07  \\
CapsNet+LOF+CTIR  &- &-  &82.90	&83.19  \\ 
ZSDNN+LOF+CTIR &-	&-	&\bf84.10	&\bf84.33  \\ \hline

\end{tabular}
}
\caption{Comparison with SOTA methods in SNIPS.}
\label{tab:sota}
\end{center}
\end{table}

\begin{table}[t]
\small
\resizebox{1.0\linewidth}{!}{
\begin{tabular}{l|cc|cc|cc|cc}
\hline
\multirow{3}{*}{Model}  &\multicolumn{4}{c}{SNIPS} &\multicolumn{4}{|c}{CLINC}   \\ \cline{2-9}
& \multicolumn{2}{c}{ZSID}  & \multicolumn{2}{|c}{GZSID}  &
\multicolumn{2}{|c}{ZSID}   &
\multicolumn{2}{|c}{GZSID}  \\ \cline{2-9} 
&Acc  &F1  &Acc  &F1  &Acc  &F1  &Acc  &F1 \\\hline

CNN + CTIR   &\bf94.73	&\bf94.73	&\bf82.31	&\bf81.03  &\bf85.11	&\bf85.20  &82.91	&78.88\\
\quad w/o MT  &94.09	&94.08	&82.24	&80.96  &84.96	&85.01  &82.85	&78.73\\
\quad w/o SS  &94.54	&94.53	&80.55	&79.60  &83.78	&83.82  &\bf84.24	&\bf80.25\\
\quad SS $\rightarrow$ ES &81.89	&80.95	&80.51	&79.50  &82.44	&82.54  &82.54	&78.29\\ \hline

CapsNet + CTIR  &\bf94.84 &\bf94.84 &\bf83.21 &\bf81.99  &\bf87.01	&\bf86.91  &\bf86.58	&\bf84.09\\
\quad w/o MT   &93.68	&93.67	&81.02	&79.07 &85.17	&85.07  &86.19	&83.64\\
\quad w/o SS &90.30	&90.26	&83.20	&81.94 &85.37	&85.16  &86.26	&83.69\\ 
\quad SS $\rightarrow$ ES   &82.54	&81.96	&83.11	&81.90 &79.97	&79.79 &86.57	&83.92 \\\hline
ZSDNN + CTIR 	&\bf95.07	&\bf95.07	&\bf85.12	&\bf84.55  &\bf93.57	&\bf93.62  &\bf84.30 &\bf83.18\\
\quad w/o MT &93.55	&93.52	&84.90	&84.30 &93.40	&93.46  &84.03	&82.81\\ \hline

CDSSM + CTIR &\bf94.14	&\bf94.14	&78.51	&\bf76.66  &\bf83.07	&\bf82.52  &\bf82.87	&\bf79.65\\ 
\quad w/o MT &93.84	&93.83	&\bf78.73	&76.01  &82.68	&82.12  &82.79	&79.57\\ \hline

\end{tabular}
}
\caption{Results of ablation study. w/o MT and w/o SS indicate removing multi-task learning and the similarity scorer, respectively. SS $\rightarrow$ ES means replacing our Similarity Scorer with the word embedding based similarity.}
\label{tab:ablation}

\end{table}

\paragraph{Comparison with SOTA methods.}
Table \ref{tab:sota} presents the comparison of the proposal with SOTA methods under their experimental settings as described in the Baselines. As we can see, +CTIR methods clearly outperform ReCapsNet in ZSID, and the +LOF+CTIR methods outperform SEG, which is also a two-stage approach, in GZSID. The +LOF method is used as our stronger baseline and its performance is comparable with SEG in GZSID. One may find that ZSDNN+LOF+CTIR is better than ZSDNN+CTIR in Table \ref{tab:sota}, while lags behind in our GZSID setting. The reason is that their dataset splitting assigns less importance to the seen intents, where the performance of +LOF is relatively poor. 
\paragraph{Ablation Study.}
We can observe from Table \ref{tab:ablation} that: 1) Our similarity scorer clearly outperforms the word embedding based intent similarity, especially in the ZSID setting. The advantage of SS over ES is less obvious in GZSID because the model trained with CTIR can directly perform GZSID, which makes the effect of the similarity matrix less significant. This can also explain for the phenomenon that w/o SS outperforms CNN+CTIR in CLINC under GZSID setting. 2) In spite of this, removing the similarity scorer has a negative impact on the model performance. 3) Multi-task learning consistently benefits the CTIR framework in all circumstances. For GZSID, the improvement directly comes from the disentangling effect of MT. For ZSID, we attribute the improvement to the well-learned relationship between seen and unseen intents, which is a side-effect of increasing the distance between unseen and seen intent representations.
We also investigate the performance variation with the increase of $\alpha$ and $\lambda^{\prime}$, please see Appendix D$^1$ for details.

\section{Related Work}
\paragraph{Zero-shot Learning.} 
In computer vision, ZSL is a well-developed sub-field, where a common approach is to relate unseen classes with seen classes through visual attributes \cite{farhadi2009describing,parikh2011relative} or word2vec representations of the class names \cite{Mikolov13,frome2013devise,socher2013zero}. 
For ZSID, transformation-based methods \cite{XiaZYCY18,liu2019reconstructing} calculate the inter-intent similarity based on the word embeddings of intent labels, and use it to transform the predictions from seen intents to unseen intents. Compatibility-based methods \cite{chen2016zero,kumar2017zero} attempt to learn a shared semantic space for label names and utterances from seen data, and then measure the similarity between a test utterance and each unseen label in this space. 
There are also studies resorting to external knowledge, e.g., label ontologies \cite{ferreira2015zero} or human-defined attributes \cite{yazdani2015model,ZhangLG19}, which, however, are laborious to obtain.

\paragraph{Generalized Zero-shot Learning.}
Socher~\shortcite{socher2013zero} and Zhang~\shortcite{zhang2016online} design different model architectures to consider the probability of each sample coming from an unseen class before final classifying.
In the task of intent detection, ReCapsNet-ZS \cite{liu2019reconstructing} enhanced CapsNet \cite{XiaZYCY18} in GZSID by modeling the correlation between the dimensions of word embeddings, which can find a more accurate connection between seen and unseen intents. On the basis of ReCapsNet-ZS, Yan~\shortcite{Guangfeng} proposed a two-stage GZSID method that combines unknown intent detection and ZSID, which successfully resolves the domain-shift problem. However, this is at the cost of the performance in seen intents, which is the majority in real-world applications.

\paragraph{Class-transductive Zero-shot Learning.} Class- transductive ZSL supports the use of semantic information (typically textual descriptions) about the unseen classes during training. In the CV field, class-transductive methods are used to infer the relationship between seen and unseen classes \cite{Zhen15,Zhenyong18} or directly predict the parameters of unseen intent classifiers \cite{Mohamed,Xiaolong}. In comparison, CTIR uses the unseen label names as training instances to learn unseen intent representations, which takes advantage of the fact that label names and utterances both come from the textual domain.

\section{Conclusions}

In this paper, we propose a class-transductive framework, CTIR, to overcome the limitations of existing ZSID models. 
CTIR utilizes the unseen label names as input utterances and includes the unseen classes into the prediction space during training. Under this framework, we present a multi-task learning objective in the training stage to encourage the model to learn the distinctions between unseen and seen intents. In the inference stage, we develop a similarity scorer, which can better associate the inter-intent connections based on the learned representations. Experiments on two benchmarks show that CTIR can bring considerable improvement to ZSID systems with different ZSL strategies and backbone networks.

\section*{Acknowledgments}
This work was supported by National Natural Science Foundation of China (No. 61976207, No. 61906187)

\bibliographystyle{named}
\bibliography{ijcai21}

\end{document}